\pdfoutput=1
\documentclass[11pt]{article}

\usepackage{EMNLP2023}

\usepackage{times}
\usepackage{latexsym}

\usepackage[T1]{fontenc}
\usepackage[utf8]{inputenc}

\usepackage{microtype}

\usepackage{inconsolata}

\usepackage{graphicx}
\usepackage{amsmath}
\usepackage{amssymb}
\usepackage{booktabs}

\usepackage{soul}
\usepackage{xspace}
\usepackage{xcolor}
\usepackage{array}
\usepackage{multirow}

\makeatletter
\DeclareRobustCommand\onedot{\futurelet\@let@token\@onedot}
\def\@onedot{\ifx\@let@token.\else.\null\fi\xspace}
\def\eg{\emph{e.g}\onedot} 
\def\ie{\emph{i.e}\onedot} 
 
\def\etc{\emph{etc}\onedot} \def\vs{\emph{vs}\onedot}

\makeatother

\definecolor{darkcyan}{RGB}{0,113,194}
\definecolor{forestgreen}{RGB}{34,139,34}
\definecolor{codegreen}{rgb}{0,0.6,0}
\definecolor{codegray}{rgb}{0.5,0.5,0.5}
\definecolor{codepurple}{rgb}{0.58,0,0.82}
\definecolor{backcolour}{rgb}{0.95,0.95,0.92}
\title{Multiple-Question Multiple-Answer Text-VQA}

\author{Peng Tang\thanks{Equal contribution.} \ \ \ Srikar Appalaraju$^{*}$ \ \ \ R. Manmatha \ \ \ Yusheng Xie \ \ \ Vijay Mahadevan \\
AWS AI Labs \\
{\tt\small\{tangpen, srikara, manmatha, yushx, vmahad\}@amazon.com}
}
\begin{document}
\maketitle
\begin{abstract}
We present Multiple-Question Multiple-Answer (MQMA), a novel approach to do text-VQA in encoder-decoder transformer models. The text-VQA task 
%(unlike generic VQA), 
requires a model to answer a question by understanding multi-modal content: text (typically from OCR) and an associated image. 
	To the best of our knowledge, almost all previous approaches for text-VQA process a single question and its associated content to predict a single answer. In order to answer multiple questions from the same image, each question and content are fed into the model multiple times. 
	In contrast, our proposed MQMA approach takes multiple questions and content as input at the encoder and predicts multiple answers
	at the decoder in an auto-regressive manner \underline{at the same time}.
	We make several novel architectural modifications to standard encoder-decoder transformers to support MQMA.
	We also propose a novel MQMA denoising pre-training task which is designed to teach the model to align and delineate multiple questions and content with associated answers.
	MQMA pre-trained model achieves state-of-the-art results on multiple text-VQA datasets, 
	each with strong baselines. Specifically, on OCR-VQA (+2.5\%), TextVQA (+1.4\%), ST-VQA (+0.6\%), DocVQA (+1.1\%) absolute improvements over the previous state-of-the-art approaches.
\end{abstract}

\section{Introduction}
\label{sec:intro}

\begin{figure}[t]
  \centering
  \includegraphics[width=1\linewidth]{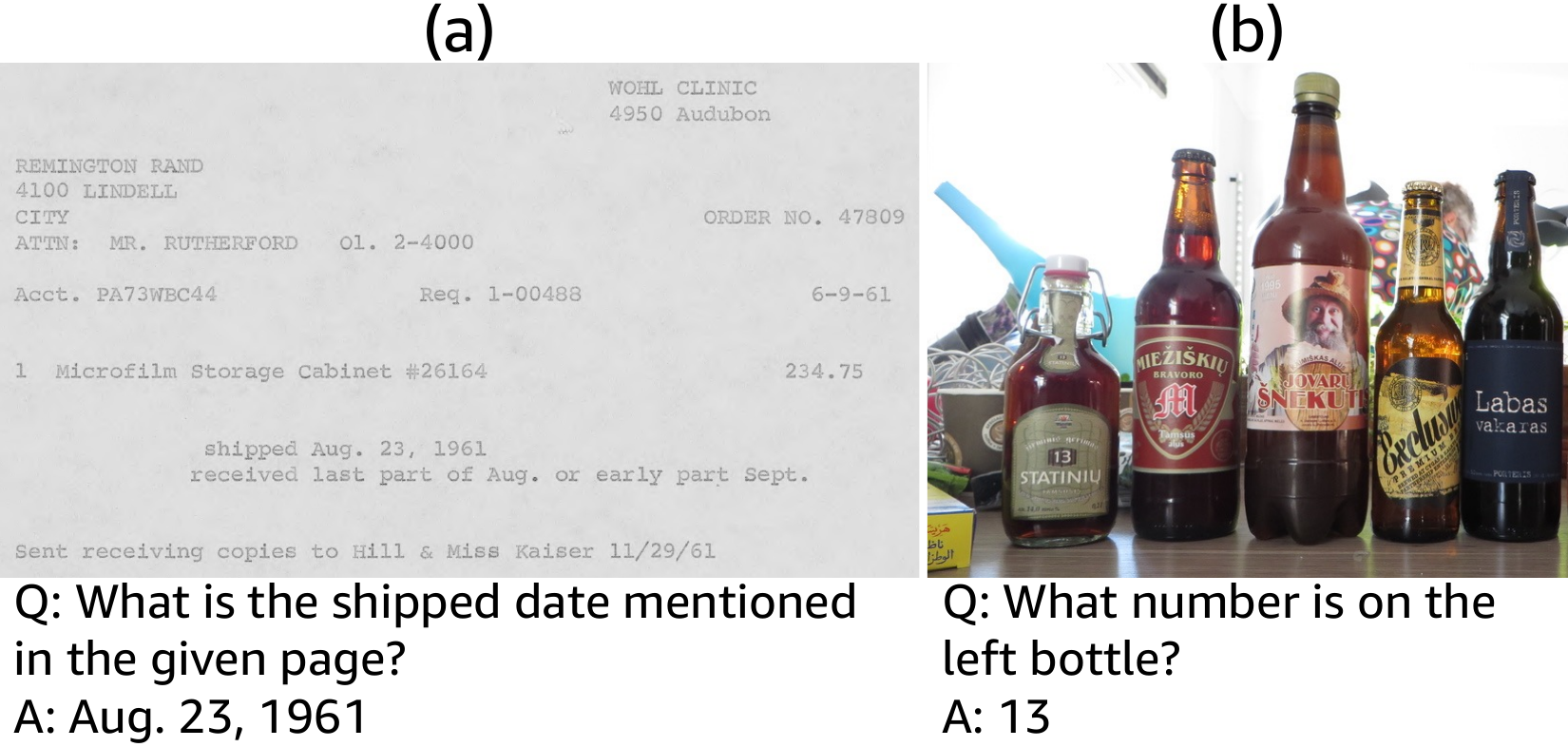}
  \caption{\textbf{Examples of text-VQA}. Examples are from (a) DocVQA \citep{mathew2021docvqa} for document VQA and (b) ST-VQA \citep{biten2019scene} for scene-text VQA. Answering questions for text-VQA requires multi-modal information, including visual, language, and layout information. Zoom in to see better.}
  \label{fig:textvqa}
\end{figure}

The task of text-based Visual Question Answering (text-VQA) requires answering questions related to a given image by understanding the text and visual contents in the image. Unlike generic VQA \citep{antol2015vqa}, where the task is to answer questions mainly using visual information,
the text-VQA task involves multiple modalities (\ie, visual, language, and layout) to answer questions \citep{biten2022latr,hu2020iterative,appalaraju2021docformer,huang2022layoutlmv3,kant2020spatially,mathew2021docvqa,mathew2020document,xu2020layoutlm,xu2021layoutlmv2,yang2021tap,appalaraju2023docformerv2}. 
The task needs a model to not only consume multiple modalities (text and image) but also to reason within and across modalities to answer a question (see Figure \ref{fig:textvqa}).

\begin{figure}[t]
    \centering
    \includegraphics[width=1\linewidth]{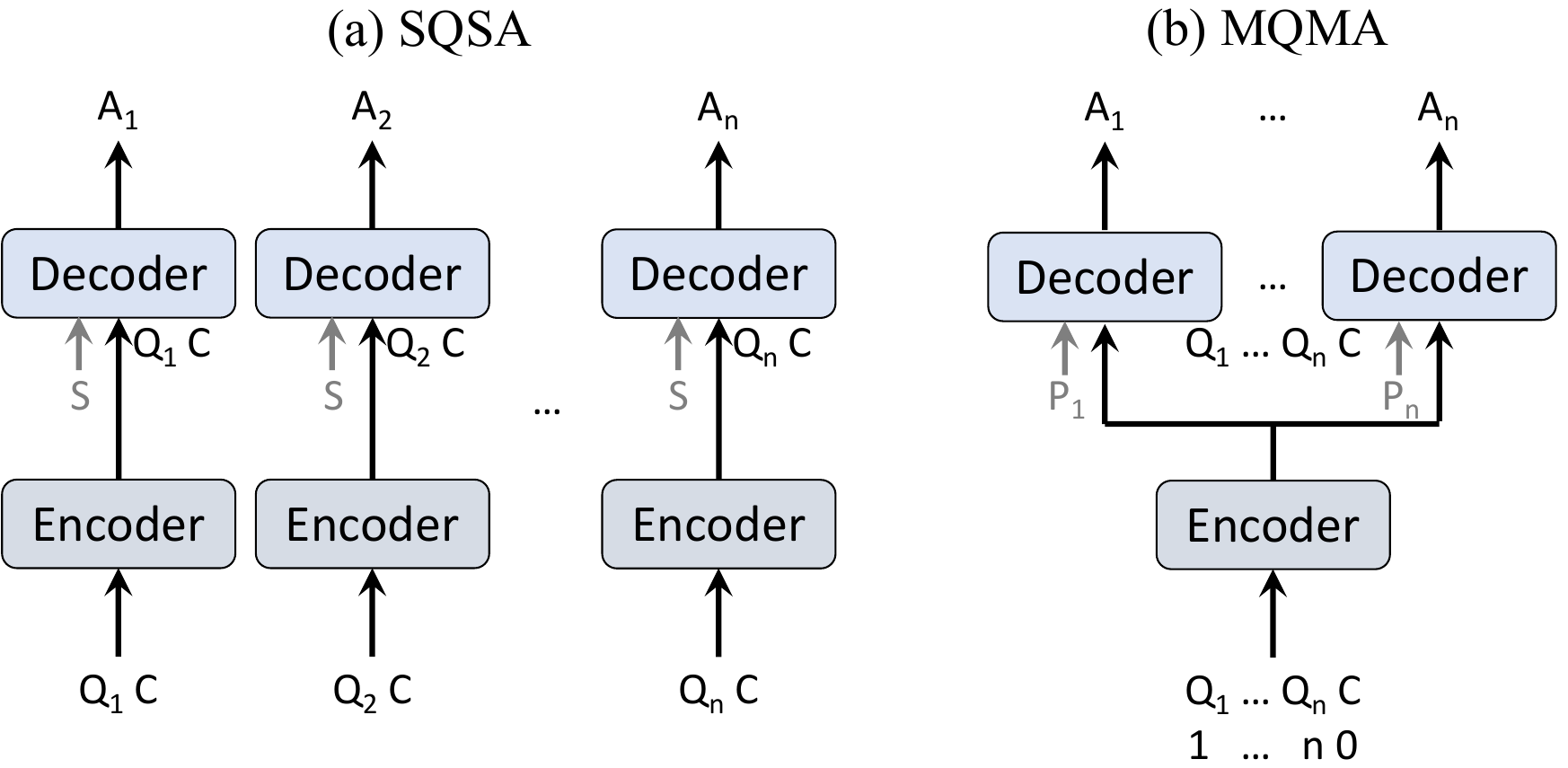}
    \caption{\textbf{Single-Question Single-Answer (SQSA) \vs Multiple-Question Multiple-Answer (MQMA)}.
    Q$_i$/A$_i$/P$_i$ ($i \in \{1, 2, ..., n\}$): the $i$-th question/answer/prompt, C: content, S: [START] token for decoder.
    $i$ ($i \in \{0, 1, 2, ..., n\}$) at the bottom of (b): question index.
    SQSA and MQMA share the same architecture of encoder and decoder except for the starting token/prompt.
    The blocks with the same color share the same weights.} 
    \label{fig::sqsa_mqma}
\end{figure}

In recent years, the text-VQA task has attracted a lot of attention \citep{biten2019scene,Li_2022_WACV,mathew2021docvqa,mathew2020document,methani2020plotqa,mishra2019ocr,singh2019towards,tanaka2021visualmrc}. Almost all text-VQA approaches known to us, consume a single question and associated content to predict a single answer. We call these approaches Single-Question Single-Answer (SQSA) text-VQA, see Figure \ref{fig::sqsa_mqma}~(a). Typical SQSA approaches 
\citep{biten2022latr,hu2020iterative,huang2022layoutlmv3,kant2020spatially,powalski2021going,xu2021layoutlmv2,yang2021tap,appalaraju2023docformerv2} first extract text in a given image using an OCR engine. Then the entire content -- image, OCR text and in some cases bounding box information 
\citep{biten2022latr,powalski2021going,appalaraju2023docformerv2}, along with the text of a single question are fed to a multi-modal transformer model which then predicts an answer.

Real-world text-VQA scenarios often involve multiple questions. For example, a user may ask multiple questions about a single image, or a group of users may ask different questions about the same image. Existing text-VQA models are not well-equipped for answering multiple questions. These models typically process a single question and its associated content to predict a single answer. In order to answer multiple questions from the same image, each question and content are fed into the model multiple times. This is inefficient and can lead to sub-optimal performance (Sec. \ref{sec::exp}). 

MQMA can address the limitations of existing text-VQA models. MQMA takes multiple questions and content as a single input sequence and predicts multiple answers \textul{at the same time}. This also opens up a possibility for the model to leverage correlations between multiple questions and content to improve accuracy.
Our choice of architecture for MQMA is an encoder-decoder seq-to-seq transformer \citep{vaswani2017attention},
see Figure \ref{fig::sqsa_mqma}~(b). 
In order to facilitate MQMA in this architecture, we introduce question index embedding at encoder and learnable prompt-based decoding, so that the model learns to align 
multiple questions and content with the respective predicted answers during auto-regressive decoding (\ie, Q1 $\rightarrow$ A1, Q2 $\rightarrow$ A2 $\dots$, \etc). During inference, each answer has its own prompt to associate the corresponding question and content
and different answers are decoded separately.
At the core of our approach is a novel MQMA unsupervised denosing pre-training task. Unlike the standard denoising language modeling task \citep{raffel2020exploring} used in the 
previous state-of-the-art text-VQA approaches \citep{biten2022latr,powalski2021going,appalaraju2023docformerv2}, our MQMA denoising task pre-trains on unlabeled document data on a proxy VQA task, \ie, a denoising language modeling task formulated as a VQA task,
to align the pre-training task and the downstream text-VQA task better. 
We highlight the contributions of our paper as follows.

\begin{itemize}
	\item To our knowledge, we are the first to propose MQMA, a novel approach to consume multiple questions and content as a single input sequence and predict multiple answers \textit{at the same time} for text-VQA (see Section \ref{sec::architecture}).
	\item We also propose an MQMA unsupervised denoising task, a novel way to train a multi-modal encoder-decoder transformer on a denoising language modeling posed as a text-VQA task (see Section \ref{sec::pretrain}).
	\item The MQMA pre-trained model achieves state-of-the-art results on the OCR-VQA, TextVQA, ST-VQA, and DocVQA datasets, each with strong baselines. In particular, +2.5\% on OCR-VQA, +1.4\% on TextVQA, +0.6\% on ST-VQA, and +1.1\% on DocVQA (see Section \ref{sec::exp}). 
\end{itemize}

\begin{figure*}[t]
    \centering
    \includegraphics[width=0.85\linewidth]{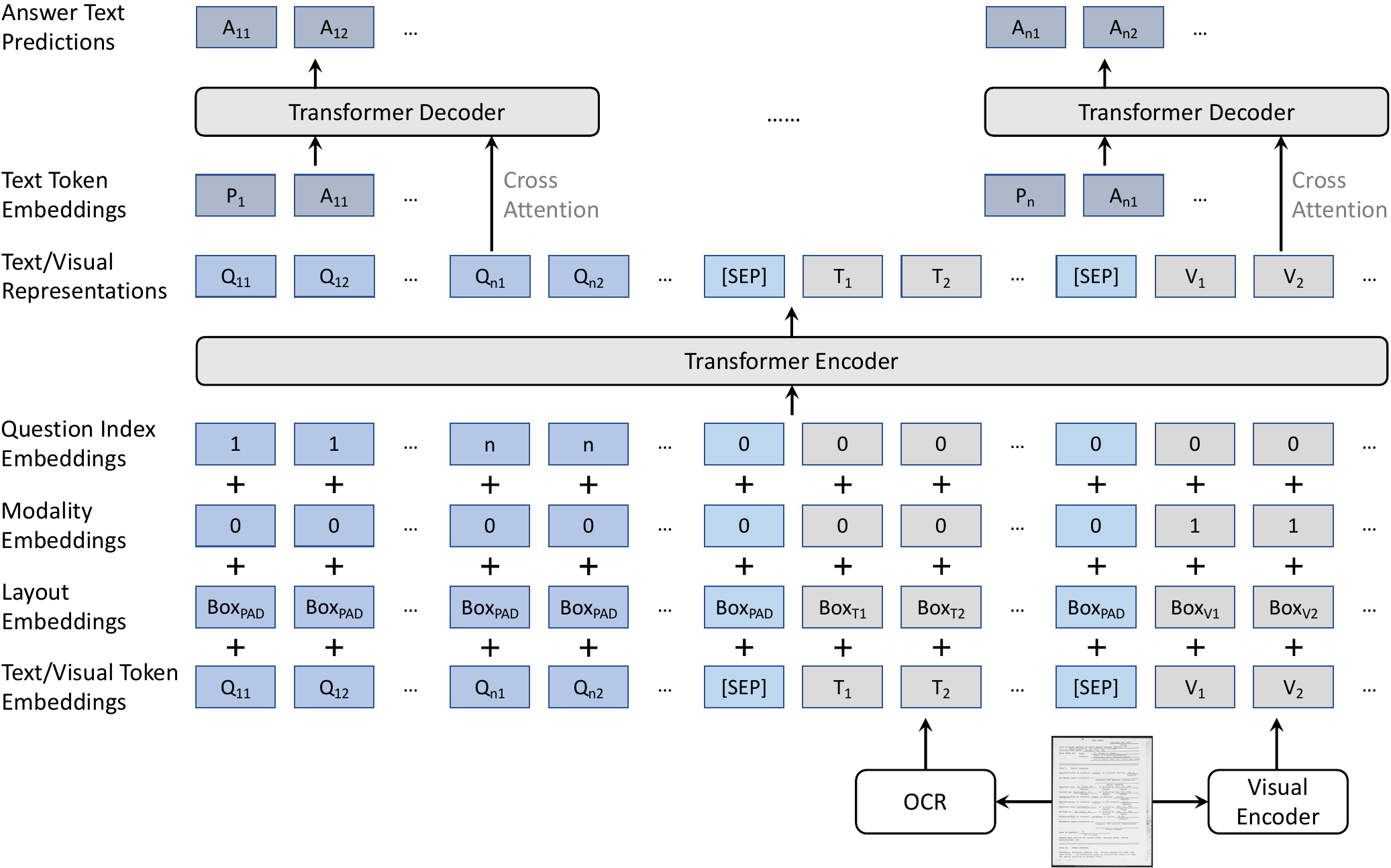}
    \caption{\textbf{MQMA Approach:} Encoder-Decoder Transformer model architecture for the proposed MQMA approach. Please note, transformer decoder has shared weights and is to be interpreted as a single decoder.}
    \label{fig::architecture}
    \vspace{-0.3cm}
\end{figure*}

\section{Related Work}
\label{sec::related_word}

Text-VQA has attracted more and more attention recently \citep{biten2019scene,kafle2018dvqa,kahou2017figureqa,mathew2022infographicvqa,mathew2021docvqa,mathew2020document,methani2020plotqa,mishra2019ocr,singh2019towards,tanaka2021visualmrc}.
Focusing on different types of images with texts, several works introduce various text-VQA datasets,
including OCR-VQA \citep{mishra2019ocr} for book and movie covers,
TextVQA \citep{singh2019towards} and ST-VQA \citep{biten2019scene} for scene-text images,
DocVQA \citep{mathew2021docvqa,mathew2020document} for document images, \etc.
Unlike generic VQA \citep{antol2015vqa} which answers questions by reasoning visual contents, 
text-VQA reasons from both text and visual contents in images to answer questions,
which introduces more challenges to the text-VQA task compared with the generic VQA.

The most common text-VQA pipeline first extracts texts and bounding boxes using OCR, and then feed multi-modal inputs (\ie, texts, bounding boxes, and image) into multi-modal models (\eg, multi-modal transformers) to get predictions \citep{biten2022latr,gao2020multi,hu2020iterative,huang2022layoutlmv3,kant2020spatially,li2021structurallm,lu2021localize,powalski2021going,xu2021layoutlmv2,yang2021tap,appalaraju2023docformerv2}.
\citet{xu2020layoutlm} propose LayoutLM based on the encoder only transformer model BERT \citep{kenton2019bert}
by using both language and layout information as inputs.
\citet{xu2021layoutlmv2} and \citet{huang2022layoutlmv3} add visual information to the inputs of LayoutLM to improve the accuracy.
\citet{hu2020iterative} and \citet{kant2020spatially} use multi-modal transformers to fuse information from different modalities and select answers from either a fixed vocabulary or OCR texts by a pointer network \citep{vinyals2015pointer}.
\citet{biten2022latr}, \citet{powalski2021going}, and \citet{appalaraju2023docformerv2} propose encoder-decoder transformer based approaches which encode multi-modal information and decode the answer in an auto-regressive manner \citep{raffel2020exploring}.
These approaches do text-VQA in a Single-Question Single-Answer (SQSA) way by answering a single question at a time.
Similar to \citep{biten2022latr,powalski2021going,appalaraju2023docformerv2}, our approach is built on top of encoder-decoder transformers.
Unlike previous approaches that answer a single question at a time,
our approach answers multiple questions at a time using our proposed Multiple-Question Multiple-Answer (MQMA) approach.
%which reduces encoder inference latency.

Before fine-tuning on text-VQA datasets, previous approaches pre-train their models on unlabeled data using tasks like masked language modeling \citep{huang2022layoutlmv3,xu2021layoutlmv2,xu2020layoutlm,yang2021tap}, image-text matching \citep{yang2021tap}, and the standard denoising \citep{biten2022latr,powalski2021going,appalaraju2023docformerv2}.
These pre-training tasks do not align well with the downstream task text-VQA,
which may limit the accuracy on the downstream task.
In contrast, we propose a new unsupervised pre-training task MQMA denoising
which pre-trains the model in a proxy VQA task.
The MQMA denoising task aligns the pre-training task with the downstream task and improves the text-VQA accuracy.

\section{MQMA Model Architecture}
\label{sec::architecture}

In this section, we discuss in detail the MQMA model architecture. Our choice of architecture for MQMA is an encoder-decoder 
transformer model (see Figure \ref{fig::architecture}). This architecture is chosen due to its popularity, versatility, and state-of-the-art text-VQA accuracy \citep{biten2022latr,powalski2021going,appalaraju2023docformerv2}.
In addition, using a vocabulary-free generative decoder lends itself as a generic 
VQA architecture over approaches which are designed for closed-vocabulary VQA \citep{antol2015vqa,wu2017visual}. 
%TODO: CITE OLD VQA PAPERS HERE.
%This then brings us to an important challenge - 
The use of decoder elicits additional challenges for MQMA as it 
is not obvious how the model can auto-regressively generate multiple answers for arbitrary number ($>1$) of input 
questions for a content.
%(image, text).

Our MQMA model is built on top of the state-of-the-art multi-modal encoder-decoder model DocFormerv2 \citep{appalaraju2023docformerv2} which is termed as the Single-Question Single-Answer (SQSA) baseline in the experiment section \ref{sec::exp}.
The input questions and content - image, OCR text, layout information 
are vectorized and fed into the transformer encoder. So the model can process multiple modalities at the same time.
See Section \ref{sec::mmei} for more details.
The transformer encoder processes these inputs with a series of self-attention layers, feed-forward layers, and 
layer normalization layers to get transformer encoder representations.
This representation is then fed into the transformer decoder,
consisting of a series of self-attention layers, cross-attention layers, feed-forward layers, and layer 
normalization layers, decoding answers as predictions in an auto-regressive manner.

In order to support MQMA functionality, the model needs to be made aware of that the input has multiple questions and that at the decoder, the model needs to appropriately align each question with the predicted 
answer. To facilitate this behavior, we introduce two key changes to the above described SQSA multi-modal 
encoder-decoder transformer architecture: \textbf{a) Question distinguishing multi-modal encoder} -  in order to distinguish 
different questions and content in the inputs, we introduce a question index embedding layer which uses 
different embeddings for different questions and content, where the embedding of index $i$ is used for the 
$i$-th question and the embedding of index $0$ is used for content (see Section \ref{sec::mmei}).
\textbf{b) Learnable prompt at the decoder} - Traditionally, a decoder is trained to auto-regressively predict a token beginning with a fixed \verb|[START]| token \citep{raffel2020exploring,vaswani2017attention}. Instead, in our approach, we introduce $n$ learnable prompts corresponding to the $n$ questions we fed into the model at the encoder. The decoder auto-regressively predicts $n$ answers beginning with these learnt prompts instead of the \verb|[START]| token. Each question uses a separate prompt to decode the corresponding answer (see Section \ref{sec::pbd}).

\subsection{Multi-modal Encoder Inputs}
\label{sec::mmei}

Both visual, language, and layout information are important to answer questions for text-VQA.
Following  common practice \citep{appalaraju2023docformerv2,biten2022latr,hu2020iterative,huang2022layoutlmv3,kant2020spatially,powalski2021going,xu2021layoutlmv2,yang2021tap},
a given input image is first processed by an OCR engine to extract text $\{\text{T}_{i}\}$ and bounding boxes $\{\text{Box}_{\text{T}i}\}$ ($i \in \{1, 2, 3, ...\}$.
The OCR text, OCR bounding boxes, question text
(Q$_{ij}, i \in \{1, 2, ..., n\}, j \in \{1, 2, ...\}$,
where $n$ corresponds to the number of questions we want to answer at a time),
and the image itself are fed into different embedding layers to get different embeddings for different modalities.
Notice that here we use text from all $n$ questions as inputs instead of a single question in previous SQSA approaches \citep{appalaraju2023docformerv2,biten2022latr,hu2020iterative,huang2022layoutlmv3,kant2020spatially,powalski2021going,xu2021layoutlmv2,yang2021tap}.
See Figure \ref{fig::architecture}.

\vspace{0.15cm}
\noindent\textbf{Text Embedding.}
We compute text embeddings for question text and OCR results.
For text, we first use the Sentence-piece tokenizer \citep{wu2016google}
to tokenize the text,
and we then use a learnable text token embedding layer to get the text token embeddings.
In particular, we add a \verb|[SEP]| token between question text tokens and OCR text tokens and append a \verb|[SEP]| token after OCR text tokens.
Apart from text token embeddings,
we compute layout embeddings of text
by using learnable layout embedding layers
to map the coordinates $(x_{1}, y_{1}, x_{2}, y_{2}, w, h)$ of text bounding boxes
into layout embeddings,
where all coordinates are normalized to [0, 1000].
For question text tokens and \verb|[SEP]|, we use a pseudo box $\verb|[BOX]|_{\text{PAD}}$ which represents the box $(0, 0, 1000, 1000, 1000, 1000)$ \citep{appalaraju2021docformer,appalaraju2023docformerv2,biten2022latr}.
We also use a learnable modality embedding layer to distinguish text modality and visual modality,
where the modality embeddings of 0 are used for the text modality.
In addition, we use a learnable question index embedding layer to distinguish different questions and content,
where the question index embeddings of $i$ and 0
are used for the $i$-th question and content respectively.
The final text embeddings are the sum of text token, layout, modality and question index embeddings.

\vspace{0.15cm}
\noindent\textbf{Visual Embedding.}
We compute visual embeddings for the image itself.
Given an input image,
first we resize the image to height 500 and width 384.
Then we split the image into 192 non-overlapped patches with size 32$\times$32.
Next we map the patches to embeddings by a linear layer with Layer Normalization \citep{ba2016layer}
and get 192 embeddings with dimension $d_{\text{emb}}$ which depends on the model size (\eg, 512 for the small size model and 768 for the base size model).
After that, we use a linear layer to map the embeddings to the final visual token embeddings 
$\{\text{V}_{i}\}_{i=1}^{128}, \text{V}_{i} \in \mathbb{R}^{d_{\text{emb}}}$,
which means the final sequence length of the visual embeddings is 128.
To compute layout embeddings of the visual part,
we first use some learnable layout embedding layers to map the location of the image patches into 192 layout embeddings,
and we then use a linear layer to map these 192 layout embeddings into the final 128 layout embeddings.
Similar to text embeddings,
the final visual embeddings are the sum of visual token embeddings, layout embeddings, modality embeddings, and question index embeddings,
where the modality embeddings of 1 and the question index embeddings of 0 are used for visual embeddings.

\subsection{Prompt-Based Decoder}
\label{sec::pbd}
In SQSA, it is straightforward to follow the standard decoding steps to do auto-regressive answer prediction beginning with the \verb|[START]| token \citep{powalski2021going,vaswani2017attention}.
For MQMA, the most naive way to get multiple answers is to decode the concatenation of multiple answers.
%However, this naive approach will result in high inference latency.
More precisely, suppose the answer sequence length is $L$,
to answer $n$ questions,
the time complexities of the self-attention layers in decoder of SQSA and MQMA are
$n \times \mathop{O}(L^{2})$ and $\mathop{O}((n \times L)^{2}) = n^{2} \times \mathop{O}(L^{2})$ respectively.
Particularly,
SQSA can decode $n$ answers in parallel which can benefit from the parallel GPU computations,
whereas MQMA has to decode $n$ answers sequentially.
All these facts show that decoding the concatenation of multiple answers for MQMA might not be a good choice. 

To address the issues mentioned above and enable parallel answer decoding for multiple-answers,
we propose a prompt-based approach for the MQMA decoder.
More precisely, we use $n$ learnable prompts $\{\text{P}_{i}\}_{i=1}^{n}$ to decode $n$ answers in parallel.
Instead of beginning with the \verb|[START]| token,
the decoder begins with the $i$-th prompt $\text{P}_{i}$ to decode the answer $\text{A}_{i}$ for the $i$-th question in an auto-regressive manner.
These prompts are learnt to associate the corresponding questions and content.
Compared with SQSA, the prompt-based MQMA decoder has almost the same decoder latency as SQSA
because the decoding processes of SQSA and MQMA are the same
except for which token the decoder begins with.
See Appendix \ref{sec_supp::time} for analyses on different MQMA approaches and why our approach is most optimal for big-oh complexity.

\section{MQMA Unsupervised Pre-training}
\label{sec::pretrain}
It is well established that pre-training followed by task specific fine-tuning almost always leads to superior performance when compared with models trained with 
just supervised fine-tuning \citep{appalaraju2021docformer,appalaraju2023docformerv2,biten2022latr,kenton2019bert,he2019moco,Chen2022PaLIAJ,hoyoro,Brown2020LanguageMA}. 
Ability to train on vast amounts of unsupervised data has a key role to play in the success of this training strategy. In language domain, a number of pre-training 
strategies inspired by cloze task \citep{Taylor1953ClozePA} have been designed, \eg, masked language modeling \citep{kenton2019bert}.
More recently, a denoising language modeling pre-training task was proposed in the T5 model \citep{raffel2020exploring} and this pre-training task has been successfully used 
in previous text-VQA models like DocFormerv2 \citep{appalaraju2023docformerv2} and LaTr \citep{biten2022latr}. The denoising language modeling task is unsupervised. The task masks spans of original text and the objective is to reconstruct the masked text during training (see ``Standard denoising'' in Table \ref{table::pretrain_task}).

\begin{table}[t]
\centering
\resizebox{\linewidth}{!}{
\begin{tabular}{l|l|l}
%   \hline
  %\toprule
  Ip / Target & Standard denoising & MQMA denoising \\
  \midrule
  Original text & \multicolumn{2}{l}{Thank you \st{for inviting} me to your party \st{last} week …} \\
  \midrule
  Input text & Thank you [MASK$_{1}$] me to your party & Q$_{1}$ Q$_{2}$ ... Q$_{n}$ [SEP] Thank you \\
  & [MASK$_{2}$] week ... & [MASK$_{1}$] me to your party [MASK$_{2}$] \\
  & & week ... \\
  \midrule
  Target & [MASK$_{1}$] for inviting [MASK$_{2}$] last ... & A$_{1}$ A$_{2}$ ... A$_{n}$ \\
  %\bottomrule
\end{tabular}
}
\caption{\textbf{Pre-training tasks}: Standard \vs MQMA denoising.}
\label{table::pretrain_task}
\end{table}

However, this standard denoising task is not well coordinated with our downstream task of text-VQA (we show in experiments, see Table \ref{table::architecture_aug}). 
%TODO: add location of experiment
In order to leverage unsupervised 
pre-training, we propose a novel MQMA denoising language modeling task as a proxy VQA task. We show that this pre-training not only helps the MQMA setting but also helps in general when the downstream task is text-VQA (see Table \ref{table::architecture_aug}). 
%TODO: add locations of the experiment
More precisely, we modify the standard denoising pre-training task to an MQMA text-VQA task by asking and answering questions on \verb|[MASK]| tokens,
see ``MQMA denoising'' Table \ref{table::pretrain_task}.
We design \verb|which| and \verb|what| style questions, \ie,\\
1) \emph{Which text tokens are masked by} $\verb|[MASK|_{i}\verb|]|$ \emph{after ``xxx''?},\\
2) \emph{What are the masked text tokens of} $\verb|[MASK|_{i}\verb|]|$ \emph{after ``xxx''?}\\
Where \verb|[MASK|$_{i}$\verb|]| corresponds to the $i$-th mask and $\textit{``xxx''}$ corresponds to the text before \verb|[MASK|$_{i}$\verb|]|.
The answer to the question above is the original text of \verb|[MASK|$_{i}$\verb|]|.
An example question-answer pair for \verb|[MASK|$_{i}$\verb|]| is\\
\emph{Q: Which text tokens are masked by} $\verb|[MASK|_{1}\verb|]|$ \emph{after ``Thank you''? - A: for inviting}

We experimentally show that this novel pre-training task is better aligned with the downstream text-VQA task and benefits the model for text-VQA even if the MQMA setting is not desired.
We also tried ``before'' style question 
formulation and found it to be not as 
beneficial when compared with the ``after'' style. 
So in experiments we stick to the ``after'' 
style questions only. 
There could be other ways to formulate the 
questions to get more benefits.

\section{Experiments}
\label{sec::exp}

\subsection{Experimental Setup}
\label{sec::exp_setup}

\noindent\textbf{Datasets and Evaluation Metrics.}
For unsupervised per-training,
we use 1M, 64M, and 64M 
unlabeled document images from the Industrial Document Library (IDL)\footnote{\url{https://www.industrydocuments.ucsf.edu/}} dataset for small, base, and large size models, respectively, following \citep{biten2022latr,appalaraju2023docformerv2}.
For text-VQA, we use OCR-VQA \citep{mishra2019ocr} for book/movie cover VQA, TextVQA \citep{singh2019towards} and ST-VQA \citep{biten2019scene} for scene-text VQA, and DocVQA \citep{mathew2021docvqa,mathew2020document} for document VQA.
See Appendix \ref{sec_supp:datasets} for more stats on these datasets.
For evaluation, we use Average Normalized Levenshtein Similarity (ANLS) \citep{biten2019icdar} which measures the similarity between predicted and ground truth answers for DocVQA and ST-VQA and the standard VQA accuracy \citep{antol2015vqa} for other datasets, following the standard evaluation protocol \citep{appalaraju2023docformerv2,biten2019scene,mathew2021docvqa,mishra2019ocr,singh2019towards}. Higher the better.

\vspace{0.15cm}
\noindent\textbf{MQMA Dynamic Data Augmentation.} During pre-training and fine-tuning, we use an MQMA specific dynamic data augmentation strategy.
%TODO: explain why dynamic
Specifically, during unsupervised pre-training, we randomly sample 5 masks at a time with uniform-random order and create 5 questions (as shown in Section \ref{sec::pretrain}).
During downstream fine-tuning, suppose we want to answer $n$ questions at a time,
%for an image, 
we randomly sample $n', n' \in \{1, 2, ..., n\}$ question-answer pairs and randomly order the $n'$ question-answer pairs.
These randomly sampled and ordered $n'$ question-answer pairs are used during fine-tuning.
So if there are $m$ questions for an image,
there will be $m^{n} + m^{n - 1} + ... + 1$ random combinations during fine-tuning. We do this to prevent any memorization and learn spurious co-relations by the model. 
During inference, we fix the order of questions and feed every $n$ questions into the model
(if the remaining number of questions is smaller than $n$ we simply feed all the remaining questions into the model).

\vspace{0.15cm}
\noindent\textbf{Implementation Details.}
Please see Appendix \ref{sec_supp::implement_details} for implementation details.

\begin{table}[t]
\centering
\resizebox{\linewidth}{!}{
% \scalebox{0.9}{
\begin{tabular}{l|c|c}
  %\toprule
  Approach & Val Accuracy (\%) & Test Accuracy (\%) \\
  \midrule
  M4C \citep{hu2020iterative} & 63.5 & 63.9 \\
  LaAP \citep{han2020finding} & 63.8 & 64.1 \\
  LaTr$_{\text{base}}$ \citep{biten2022latr} & 67.5 & 67.9 \\
  GIT \citep{wang2022git} & 67.8 & 68.1 \\
  SQSA$_{\text{base}}$ \citep{appalaraju2023docformerv2} & 69.7 & 70.3 \\
  SQSA$_{\text{large}}$ \citep{appalaraju2023docformerv2} & 71.1 & \underline{71.5} \\
  \midrule
  MQMA$_{\text{base}}$ (ours) & 71.9 & 72.4 \\
  MQMA$_{\text{large}}$ (ours) & \textbf{73.6} & \textbf{74.0} \textcolor{forestgreen}{(+2.5)} \\
  %\bottomrule
\end{tabular}
}
\caption{\textbf{Comparison on OCR-VQA} \citep{mishra2019ocr}: We answer 5 questions at a time for MQMA.  
\textcolor{forestgreen}{+2.5\%} is absolute improvement from the previous state of the art \cite{appalaraju2023docformerv2} in that class.
\textbf{Bold} indicates best and \underline{underline} indicates the previous state of the art. 
}
\label{table::ocrvqa}
\vspace{-0.4cm}
\end{table}

\begin{table}[t]
\centering
\resizebox{\linewidth}{!}{
% \scalebox{0.9}{
\begin{tabular}{l|c|c}
  %\toprule
  Approach & Val Accuracy (\%) & Test Accuracy (\%) \\
  \midrule
  LaAP \citep{han2020finding} & 41.0 & 41.4 \\
  SA-M4C \citep{kant2020spatially} & 45.4 & 44.6 \\
  SMA \cite{gao2021structured} & 44.5 & 45.5 \\
  M4C \citep{hu2020iterative} & 47.8 & - \\
  LOGOS \citep{lu2021localize} & 51.5 & 51.1 \\
  TAP + TAG \citep{wang2022tag} & 53.6 & 53.7 \\
  TAP \citep{yang2021tap} & 54.7 & 54.0 \\
  PreSTU \citep{kil2022prestu} & 56.7 & 56.3 \\
%   LaTr$_{\text{base}}$ \citep{biten2022latr} & 58.0 & 58.9 \\
  GIT$^{\dagger}$ \citep{wang2022git} & 59.9 & 59.8 \\
  LaTr$_{\text{base}}^{\dagger}$ \citep{biten2022latr} & 59.5 & 59.6 \\
  LaTr$_{\text{large}}^{\dagger}$ \citep{biten2022latr} & 61.1 & 61.6 \\
  SQSA$_{\text{base}}^{\dagger}$ \citep{appalaraju2023docformerv2} & 61.6 & 60.0 \\
  SQSA$_{\text{large}}^{\dagger}$ \citep{appalaraju2023docformerv2} & 65.6 & \underline{64.0} \\
  \midrule
  MQMA$_{\text{base}}^{\dagger}$ (ours) & 63.1 & 62.3 \\
  MQMA$_{\text{large}}^{\dagger}$ (ours) & \textbf{66.6} & \textbf{65.4} \textcolor{forestgreen}{(+1.4)} \\
  %\bottomrule
\end{tabular}
}
\caption{\textbf{Comparison on TextVQA} \citep{singh2019towards}: We answer 2 questions at a time for MQMA. $^{\dagger}$ indicates using the combination of the ST-VQA and TextVQA training sets to train the model. 
}
\label{table::textvqa}
\vspace{-0.5cm}
\end{table}

\begin{table}[t]
\centering
\resizebox{\linewidth}{!}{
% \scalebox{0.9}{
\begin{tabular}{l|c|c}
  %\toprule
  Approach & Val ANLS (\%) & Test ANLS (\%) \\
  \midrule
  M4C \citep{hu2020iterative} & 47.2 & 46.2 \\
  LaAP \citep{han2020finding} & 49.7 & 48.5 \\
  SA-M4C \citep{kant2020spatially} & 51.2 & 50.4 \\
  LOGOS \citep{lu2021localize} & 58.1 & 57.9 \\
  TAP \citep{yang2021tap} & 59.8 & 59.7 \\
  TAP + TAG \citep{wang2022tag} & 62.0 & 60.2 \\
  PreSTU \citep{kil2022prestu} & - & 65.5 \\
  LaTr$_{\text{base}}^{\dagger}$ \citep{biten2022latr} & 68.3 & 68.4 \\
  LaTr$_{\text{large}}^{\dagger}$ \citep{biten2022latr} & 70.2 & 69.6 \\
  GIT$^{\dagger}$ \cite{wang2022git} & 69.1 & 69.6 \\
  SQSA$_{\text{base}}^{\dagger}$ \citep{appalaraju2023docformerv2} & 70.1 & 68.4 \\
  SQSA$_{\text{large}}^{\dagger}$ \citep{appalaraju2023docformerv2} & 72.9 & \underline{71.8} \\
  \midrule
  MQMA$_{\text{base}}^{\dagger}$ (ours) & 70.6 & 70.0 \\
  MQMA$_{\text{large}}^{\dagger}$ (ours) & \textbf{73.9} & \textbf{72.4} \textcolor{forestgreen}{(+0.6)} \\
  %\bottomrule
\end{tabular}
}
\caption{\textbf{Comparison on ST-VQA} \citep{biten2019scene}: 
We answer 2 questions at a time for MQMA. $^{\dagger}$ indicates using the combination of the ST-VQA and TextVQA training sets to train the model.
}
\label{table::stvqa}
\vspace{-0.3cm}
\end{table}

\subsection{Comparisons with State of the Art}
\label{sec::comparisons}

% \paragraph{Results on OCR-VQA.}
\noindent\textbf{Results on OCR-VQA.}
Table \ref{table::ocrvqa} shows results of different approaches on the OCR-VQA \citep{mishra2019ocr} dataset.
Here we train our model on the training set.
We answer 5 questions at a time for MQMA (\ie, $n = 5$) because the accuracy of using different numbers of questions is similar on OCR-VQA (see Table \ref{table_supp::ocrvqa} in Appendix).
On OCR-VQA, there could be potential information leak from the questions ``Is this book related to xxx?'' to the answer of the questions ``What type of book is this?'' / ``What is the genre of this book?'' if we ask these questions together.
To avoid such information leak, we keep these two sets of questions separate and answer them separately.
See Appendix \ref{sec_supp::infoleak} for more detailed analyses.
On the OCR-VQA testing set, our MQMA approach obtains accuracy 74.0\%
which is 2.5\% higher than 71.5\% of the previous state-of-the-art SQSA approach \citep{appalaraju2023docformerv2} using the large size model.

\begin{table}[t]
    \centering
    \resizebox{\linewidth}{!}{
    % \scalebox{0.9}{
    \begin{tabular}{l|c}
        %\toprule
        Approach & Test ANLS (\%) \\
        \midrule
        LayoutLMv2$_{\text{base}}$ \citep{xu2021layoutlmv2} & 78.1 \\
        LayoutLMv2$_{\text{large}}$ \citep{xu2021layoutlmv2} & 85.3 \\
        LayoutLMv3$_{\text{base}}$ \citep{huang2022layoutlmv3} &  78.8 \\
        LayoutLMv3$_{\text{large}}$ \citep{huang2022layoutlmv3} &  83.4 \\
        StructuralLM$_{\text{large}}$ \citep{li2021structurallm} & 83.9 \\
        UDOP$_{\text{large}}$ \cite{tang2023unifying} & 84.7 \\
        ERNIE-Layout$_{\text{large}}$ \citep{peng2022ernie} & 84.9 \\
        TILT$_{\text{base}}^{\dagger}$ \citep{powalski2021going} & 83.9 \\
        TILT$_{\text{large}}^{\dagger}$ \citep{powalski2021going} & 87.1 \\
        SQSA$_{\text{base}}$ \citep{appalaraju2023docformerv2} & 83.4 \\
        SQSA$_{\text{large}}$ \citep{appalaraju2023docformerv2} & \underline{87.2} \\
        \midrule
        % \textcolor{codegray}{Pali-X \citep{chen2023pali}} & \textcolor{codegray}{86.8} \\
        \textcolor{codegray}{ERNIE-Layout$_\text{ens}$} \citep{peng2022ernie} & \textcolor{gray}{88.4} \\
        \textcolor{codegray}{GPT4} & \textcolor{codegray}{88.4}\\
        \midrule
        MQMA$_{\text{base}}$ (ours) & \textbf{84.8} \\
        MQMA$_{\text{large}}$ (ours) & \textbf{88.3} \textcolor{forestgreen}{(+1.1)} \\
    %\bottomrule
    \end{tabular}}
    \caption{\textbf{Comparison on DocVQA} \citep{mathew2021docvqa}: 
    We answer 2 questions at a time for MQMA.
    $^{\dagger}$ indicates using more QA datasets instead of only DocVQA to train the model. \textcolor{codegray}{ERNIE-Layout$_\text{ens}$} is the ensemble of 30 models and \textcolor{codegray}{GPT4} has billions of parameters, both of which are much bigger than MQMA$_{\text{large}}$ using a single model with 750M parameters.
    }
    \label{table::docvqa}
    \vspace{-0.3cm}
\end{table}

\begin{table}[t]
\centering
\resizebox{\linewidth}{!}{
% \scalebox{0.9}{
\begin{tabular}{l|c|c|c}
  %\toprule
  Approach & Data Aug. & \# Questions & ANLS \\
  \midrule
  SQSA${_\text{small}}$ & - & 1 & 73.0 \\
  \midrule
  MQMA$_{\text{small}}$ (naive) & Static & 2 & 68.6 \\
  MQMA$_{\text{small}}$ (naive) & Dynamic & 2 & 72.3 \\
  MQMA$_{\text{small}}$ (ours w/o QIE) & Dynamic & 2 & 72.7 \\ 
  MQMA$_{\text{small}}$ (ours) & Dynamic & 2 & 72.9 \\ 
  MQMA$_{\text{small}}$ (ours) + MQMA denoising & Dynamic & 2 & \textbf{74.3} \\
  MQMA$_{\text{small}}$ (ours) + MQMA denoising + FDPF & Dynamic & 2 & 74.1 \\
  %\bottomrule
\end{tabular}
}
% % \vspace{-0.2cm}
\caption{\textbf{MQMA Ablations:} Results of different MQMA architectures, training data augmentation strategies,
and pre-training tasks on the DocVQA validation set.
``MQMA$_{\text{small}}$ (naive)'' means the naive approach that concatenates answers of multiple questions to form a single long output sequence.
``MQMA$_{\text{small}}$ (ours w/o QIE)'' means our approach w/o question index embeddings.
``MQMA$_{\text{small}}$ (ours)'' means our approach.
``MQMA$_{\text{small}}$ (ours) + MQMA denoising'' means using MQMA denoising during pre-training (otherwise using standard denoising).
``MQMA$_{\text{small}}$ (ours) + MQMA denoising + FDPF'' is the same as ``MQMA$_{\text{small}}$ (ours) + MQMA denoising'' except for freezing decoder prompts during fine-tuning.
``Static'' means that we do static data generation by fixing question-answer pair combinations during training.
``Dynamic'' means that we do dynamic data generation by randomly sampling and ordering question-answer pairs during training.
}
\label{table::architecture_aug}
\vspace{-0.3cm}
\end{table}

% \paragraph{Results on ST-VQA and TextVQA.}
\vspace{0.15cm}
\noindent\textbf{Results on TextVQA and ST-VQA.}
Following previous approaches \citep{biten2022latr,appalaraju2023docformerv2}, 
we train our models on the combination of TextVQA \cite{singh2019towards} and ST-VQA \cite{biten2019scene} training sets.
We answer 2 questions at a time for MQMA (\ie, $n = 2$) because most images in TextVQA and ST-VQA only have 1 or 2 questions.
From the results shown in  Table \ref{table::textvqa} and Table \ref{table::stvqa}, our MQMA approach consistently gives the best accuracy on both datasets under different settings.
In particular, Table \ref{table::textvqa} shows that our MQMA approach obtains accuracy 65.4\% on the TextVQA testing set, which is 1.4\% higher than the previous state-of-the-art SQSA approach \citep{appalaraju2023docformerv2}.
In addition, on the ST-VQA testing set, our MQMA approach improves ANLS from 71.8\% to 72.4\% compared with the state-of-the-art SQSA approach \citep{appalaraju2023docformerv2}, see Table \ref{table::stvqa}.

\begin{figure}
    \centering
    \includegraphics[width=\linewidth]{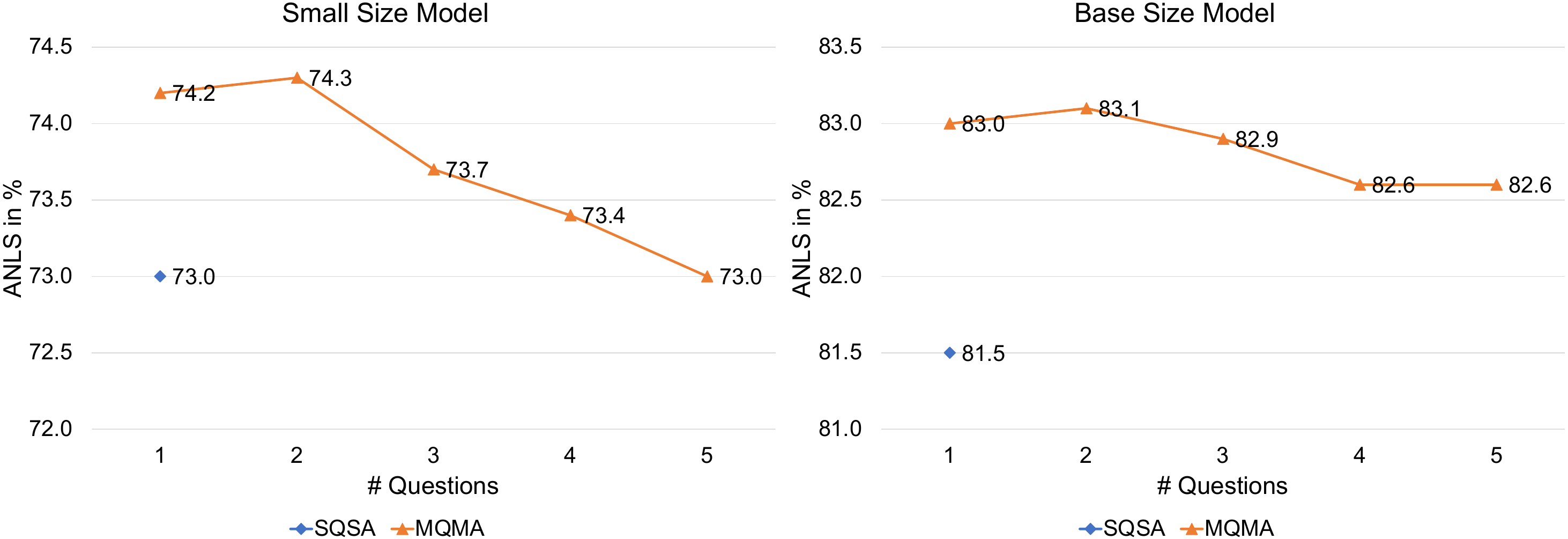}
    \caption{\textbf{MQMA Ablations}: 
    %Latency, Model Size, # Questions vs Performance (ANLS). 
    Results of different numbers of questions on the DocVQA validation set using the small size and base size models.
    We use the standard denoising task and the MQMA denoising task for SQSA and MQMA pre-training respectively.} 
    \label{fig::number_questions}
    \vspace{-0.3cm}
\end{figure}

% \paragraph{Results on DocVQA.}
\vspace{0.15cm}
\noindent\textbf{Results on DocVQA.}
Here we compare our approach with the previous state of the art on the DocVQA dataset \citep{mathew2021docvqa}.
We train our model on the combination of training and validation set and show results on the testing set (by submitting to leaderboard).
We answer 2 questions at a time for MQMA (\ie, $n = 2$) because $n = 2$ gives the best accuracy on DocVQA (see Figure \ref{fig::number_questions}).
As shown in Table \ref{table::docvqa}, our approach obtains ANLS 88.3\% on the DocVQA testing set, 1.1\% higher than 87.2\% of the previous state-of-the-art SQSA approach \citep{appalaraju2023docformerv2}.

\begin{table}[t]
\centering
\resizebox{\linewidth}{!}{
\begin{tabular}{l|c|c|c|c}
  %\toprule
  Approach & \# Questions & ANLS (\%) & ANLS of Q1 (\%) & ANLS of Q2 (\%) \\
  \midrule
  MQMA$_{\text{small}}$ & 2 & 74.3 & 75.3 & 73.6  \\
  MQMA$_{\text{small}}$ (reversed order) & 2 & 74.2 & 73.4 & 75.2 \\
  %\bottomrule
\end{tabular}
}
\caption{
%Results of different question orders 
\textbf{MQMA Ablations:} Results of different question orders on the DocVQA validation set.
The Q1/Q2 for MQMA$_{\text{small}}$ corresponds to Q2/Q1 for MQMA$_{\text{small}}$ (reversed order).
}
\label{table::q_order}
\vspace{-0.4cm}
\end{table}

\subsection{Ablation Studies}
\label{exp::docvqa}

We conduct several ablations on the DocVQA validation set to analyze the influence of different components of our approach,
including the MQMA architecture,
the training data augmentation strategy,
the unsupervised pre-training task,
the question order,
and the number of questions.
% and the model size.
If not specified, all experiments here are based on MQMA$_{\text{small}}$.

% \paragraph{The Influence of the MQMA Architecture.}
\vspace{0.15cm}
\noindent\textbf{The Influence of the MQMA Architecture.}
As we discussed in Section \ref{sec::pbd},
apart from the prompt-based decoder, we can also use a naive approach that concatenates the answers of multiple questions to form a single long output sequence.
In addition, we also remove the question index embeddings to check the influence of the question index embeddings.
Here we compare these three different MQMA architectures.
We do 2 questions 2 answers document VQA (\ie, $n=2$).
As shown in Table \ref{table::architecture_aug},
our approach obtains higher ANLS than the naive approach.
In addition, our approach has lower latency than the naive approach, see Table \ref{table_supp::approach_comparison} in Appendix. 
Adding question index embeddings also contributes to higher ANLS
because the question index embeddings help the model distinguish different questions and content.

% \paragraph{MQMA Training Data Augmentation Strategy.}
\vspace{0.15cm}
\noindent\textbf{MQMA Training Data Augmentation Strategy.}
As mentioned in Section \ref{sec::exp_setup}.
we use a dynamic training data augmentation strategy by randomly sampling and ordering question-answer pairs.
Here we compare the dynamic training data augmentation strategy with the static training data generation approach
which fixes question-answer pair combinations during training.
From Table \ref{table::architecture_aug},
we can see that using the dynamic approach obtains 3.7\% higher ANLS than the static approach.
%We use dynamic training data augmentation in the rest of our paper.

% \paragraph{The Influence of the Unsupervised Pre-training Task.}
\vspace{0.15cm}
\noindent\textbf{The Influence of the Unsupervised Pre-training Task.}
Here we study the influence of different unsupervised pre-training tasks.
From Table \ref{table::architecture_aug},
we can see that adding the MQMA denoising pre-training task improves ANLS by 1.4\% when $n = 2$.
With the new pre-training task, our MQMA approach obtains 1.3\% higher ANLS compared with SQSA.
In addition, from Figure \ref{fig::number_questions},
we can see when pre-trained with the MQMA denoising task,
even $n = 1$ contributes to higher ANLS than the SQSA baseline with the standard denoising task.
These results confirm that MQMA denoising is beneficial for text-VQA even if $n = 1$.
Also, even freezing the decoder prompts during fine-tuning obtains an ANLS of 74.1\% (\vs 74.3\%), which confirms that our pre-training task can learn good decoder prompts to associate the corresponding questions and content even without fine-tuning learnable decoder prompts.

% \paragraph{The Influence of the Question Order.}
\vspace{0.15cm}
\noindent\textbf{The Influence of the Question Order.}
In our approach, questions are concatenated with fixed order during inference.
Here we study the influence of the question order. 
From Table \ref{table::q_order}, we can see our approach is robust to the order of the questions.
This is because our model is trained with dynamic data augmentation which randomly samples and orders questions during training.

% \paragraph{The Influence of the Number of Questions.}
%\noindent\textbf{The Influence of the Number of Questions}.
\vspace{0.15cm}
\noindent\textbf{The Influence of the Number of Questions.}
We discuss the results of different numbers of questions we answer at a time (\ie, different $n$).
As we can see from Figure \ref{fig::number_questions}, our MQMA obtains higher accuracy than SQSA for $n=1$ to $5$.
Answering 2 questions at a time gives the best accuracy on DocVQA, so we use $n=2$ in Section \ref{sec::comparisons}.
See Appendix \ref{sec_supp::n_q} for the influence of the number of questions on other datasets.

\section{Conclusion}
\label{sec::conclusion}

In this paper, we propose a Multiple-Question Multiple-Answer (MQMA) text-VQA approach.
Unlike previous approaches that process a single question each time,
MQMA can answer multiple questions at a time.
In addition, we propose an MQMA denoising task for unsupervised pre-training.
The MQMA denoising task aligns the pre-training task with the downstream text-VQA task to improve accuracy.
Experimental results show that
the proposed approach improves accuracy 
on a variety of challenging text-VQA datasets compared
with the previous state of the art.

\bibliography{anthology,custom}
\bibliographystyle{acl_natbib}

\appendix

\section{Time Complexity and Latency of SQSA and Different MQMA Approaches}
\label{sec_supp::time}
\begin{figure}[t]
    \centering
    \includegraphics[width=\linewidth]{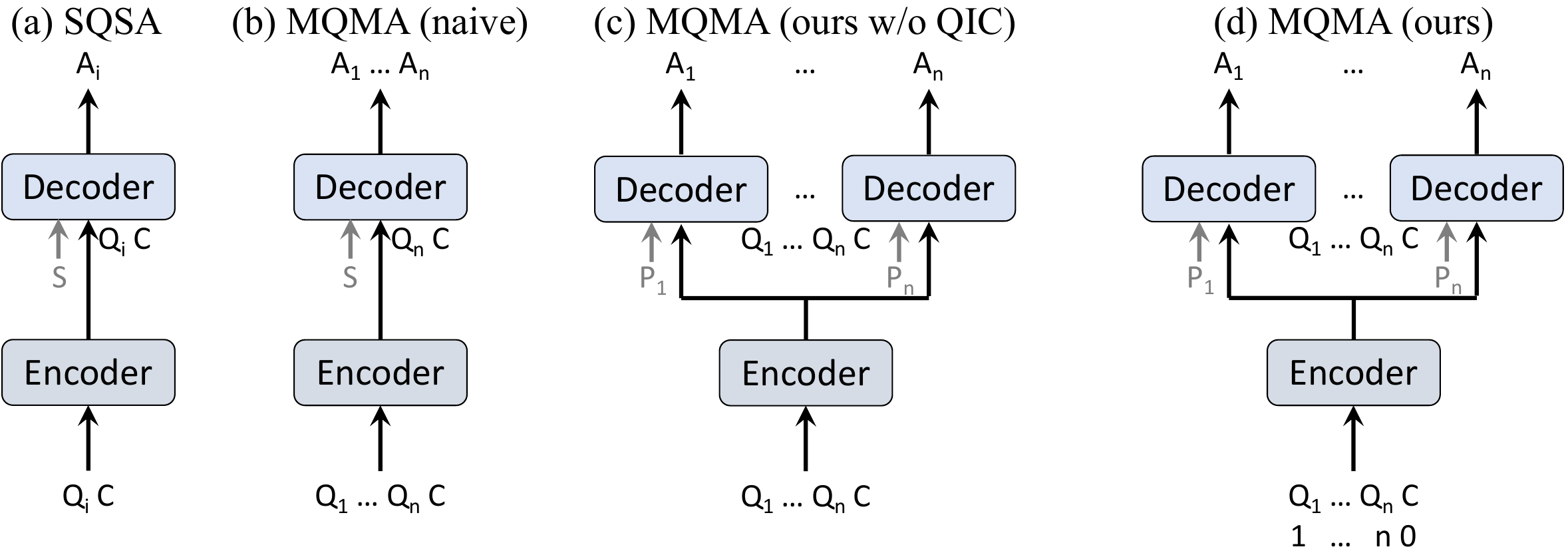}
    \caption{\textbf{Architecture Comparisons among SQSA and Different MQMA approaches:}
    SQSA: the SQSA baseline,
    MQMA (naive): the naive MQMA approach that concatenates answers of multiple questions to form a single long output sequence,
    MQMA (ours w/o QIE): our MQMA approach w/o question index embeddings,
    MQMA (ours): our MQMA approach,
    Q$_i$/A$_i$/P$_i$ ($i \in \{1, 2, ..., n\}$): the $i$-th question/answer/prompt, C: content, S: [START] token for decoder.
    $i$ ($i \in \{0, 1, 2, ..., n\}$) at the bottom of (d): question index.}
    \label{fig_supp::approach_comparison}
\end{figure}

We do detailed time complexity and latency analyses of SQSA and different MQMA approaches here. See Figure \ref{fig_supp::approach_comparison} for the architectures of SQSA and different MQMA approaches.
Suppose we have $n$ questions, the sequence length of each question is $L_{\text{Q}}$, the sequence length of content is $L_{\text{C}}$, and the sequence length of each answer is $L_{\text{A}}$. Without loss of generality, $L_{\text{Q}} << L_{\text{C}}$.

For SQSA, to answer each question, the time complexity of each self-attention layer in the encoder is $O\left((L_{\text{Q}} + L_{\text{C}})^{2}\right) \approx O\left(L_{\text{C}}^{2}\right)$.
The time complexity of each self-attention layer and cross-attention layer in the decoder is 
$O\left(L_{\text{A}}^{2} + L_{\text{A}} * (L_{\text{Q}} + L_{\text{C}})\right) \approx O\left(L_{\text{A}}^{2} + L_{\text{A}} * L_{\text{C}}\right)$, where $L_{\text{A}}^{2}$ is from the self-attention layer and $L_{\text{A}} * L_{\text{C}}$ is from the cross-attention layer.
So the encoder and decoder time complexities of answering $n$ questions are $n * O\left(L_{\text{C}}^{2}\right)$ and $n * O\left(L_{\text{A}}^{2} + L_{\text{A}} * L_{\text{C}}\right)$ respectively.

For MQMA (naive), we answer $n$ questions at a time.
The time complexity of each self-attention layer in the encoder to answer $n$ questions is $O\left((n * L_{\text{Q}} + L_{\text{C}})^{2}\right) \approx O\left(L_{\text{C}}^{2}\right)$ ($n * L_{\text{Q}} << L_{\text{C}}$) which is $\frac{1}{n}$ of the encoder time complexity of SQSA.
The time complexity of each self-attention layer and cross-attention layer in the decoder to answer $n$ questions is
$O\left((n * L_{\text{A}})^{2} + (n * L_{\text{A}}) * (L_{\text{Q}} + L_{\text{C}})\right) \approx n * O\left(n * L_{\text{A}}^{2} + L_{\text{A}} * L_{\text{C}}\right)$ which is higher than the decoder time complexity $n * O\left(L_{\text{A}}^{2} + L_{\text{A}} * L_{\text{C}}\right)$ of SQSA.

For MQMA (ours w/o QIE) and MQMA (ours), we answer $n$ questions at a time.
The time complexity of each self-attention layer in the encoder to answer $n$ questions is the same as MQMA (naive) because the input sequence length of different MQMA approaches is the same.
The time complexity of each self-attention layer and cross-attention layer in the decoder to answer $n$ questions is the same as SQSA because we decode $n$ answers separately as in SQSA.

\begin{table*}[t]
\centering
\resizebox{\linewidth}{!}{
\begin{tabular}{l|l|l|l|l}
%   \hline
  \toprule
   & SQSA & MQMA (naive) & MQMA (ours w/o QIE) & MQMA (ours) \\
  \midrule
  Encoder Time Complexity & $\textcolor{red}{n\ *\ } O\left(L_{\text{C}}^{2}\right)$ & $O\left(L_{\text{C}}^{2}\right)$ & $O\left(L_{\text{C}}^{2}\right)$ & $O\left(L_{\text{C}}^{2}\right)$ \\
  Encoder Latency (ms/image) & {\color{red}19.7} & \textbf{11.5} & \textbf{11.5} & \textbf{11.5} \\
  \hline
  Decoder Time Complexity & $n * O\left(L_{\text{A}}^{2} + L_{\text{A}} * L_{\text{C}}\right)$ & $n * O\left(\textcolor{red}{n\ *\ } L_{\text{A}}^{2} + L_{\text{A}} * L_{\text{C}}\right)$ & $n * O\left(L_{\text{A}}^{2} + L_{\text{A}} * L_{\text{C}}\right)$ & $n * O\left(L_{\text{A}}^{2} + L_{\text{A}} * L_{\text{C}}\right)$ \\
  Decoder Latency (ms/image) & \textbf{68.9} & {\color{red}77.6} & \textbf{68.9} & \textbf{68.9} \\
  \bottomrule
\end{tabular}
}
\caption{\textbf{Time Complexity and Latency Comparisons among SQSA and Different MQMA Approaches:}
    SQSA: the SQSA baseline,
    MQMA (naive): the naive MQMA approach that concatenates answers of multiple questions to form a single long output sequence,
    MQMA (ours w/o QIE): our MQMA approach w/o question index embeddings,
    MQMA (ours): our MQMA approach,
    $n$: the number of questions,
    $L_{\text{C}}$: the sequence length of content,
    $L_{\text{A}}$: the sequence length of answer.
    The latency numbers here are from MQMA$_{\text{small}}$ on DocVQA \cite{mathew2021docvqa}.}
\label{table_supp::approach_comparison}
\end{table*}

We summarize the time complexities of different approaches and report latency in Table \ref{table_supp::approach_comparison}.
Our MQMA approaches give lower encoder time complexity and latency than SQSA.
In addition, the decoder time complexity and latency of MQMA (ours w/o QIE) and MQMA (ours) are the same as that of SQSA and are lower than that of MQMA (naive).
So MQMA (ours w/o QIE) and MQMA (ours) give the lowest overall time complexity and latency among all these approaches.

\section{Datasets}
\label{sec_supp:datasets}

\begin{table}[t]
\centering
\resizebox{\linewidth}{!}{
\begin{tabular}{l|c|c|c}
%   \hline
  \toprule
  Dataset & Train Set & Val Set & Test Set\\
  \midrule
  OCR-VQA \citep{mishra2019ocr} & 166K/801.7K & 20.7K/100K & 20.8K/100.4K \\
  TextVQA \citep{singh2019towards} & 21.9K/34.6K & 3.2K/5K & 3.3K/5.7K \\
  ST-VQA \citep{biten2019scene} & 17K/23.4K & 1.9K/2.6K & 3K/4.1K \\
  DocVQA \citep{mathew2020document,mathew2021docvqa} & 10.2K/39.5K & 1.3K/5.3K & 1.3K/5.2K \\
  \bottomrule
\end{tabular}
}
\caption{\textbf{Dataset Stats:} The number of images/questions of different text-VQA datasets.}
\label{table_supp::datasets}
\end{table}

As stated in the main paper, we use OCR-VQA \citep{mishra2019ocr} for book/movie cover VQA, TextVQA \citep{singh2019towards} and ST-VQA \citep{biten2019scene} for scene-text VQA, and DocVQA \citep{mathew2021docvqa,mathew2020document} for document VQA.
See Table \ref{table_supp::datasets} for details of these text-VQA datasets.
As we can see, there are on average $\sim 5$ questions/image on OCR-VQA, 1 or 2 questions/image on TextVQA and ST-VQA, and on average $\sim 4$ questions/image on DocVQA.

\section{Implementation Details}
\label{sec_supp::implement_details}
\vspace{0.15cm}
\noindent\textbf{Pre-training.} We use small, base, and large size models which are termed as MQMA$_{\text{small}}$, MQMA$_{\text{base}}$, and MQMA$_{\text{large}}$, respectively.
Our model is first initialized from the T5 pre-trained weights \citep{raffel2020exploring},
then pre-trained on the unlabeled document data following DocFormerv2 \citep{appalaraju2023docformerv2} - we call this model 
%DFv2 and is used 
as SQSA basline in our experiments. 
%DFv2 
SQSA is next pre-trained on the same unlabeled document data using the MQMA denoising task descried in Sec. 4 of the main paper. In both, we pre-train for 50/3/3 epochs on 1M/64M/64M IDL data for the small/base/large size model. 
We also do not do any text augmentation \citep{ma2019nlpaug,Feng2021ASO} or multi-modal augmentation \citep{Hao2022MixGenAN}. We simply normalize the images to unit mean and variance for training stability. 
The maximum input sequence length of the text token embeddings is set to 512.
The input sequence length of the visual token embeddings is set to 128.
The learnable prompt P$_{i}$ is first initialized by the embeddings of ``answer of question $i$:''.

\vspace{0.15cm}
\noindent\textbf{Fine-Tuning.}
For text-VQA fine-tuning, we train our models for 8 epochs on OCR-VQA and for 50 epochs on other datasets.
The learning rate is set to 0.0001 and the AdamW \citep{loshchilov2018decoupled} optimizer is used to train our models.
Our training batch size is set to 128.
The maximum input sequence length of the text token embeddings is set to 2048 for small and base size models and 1024 for large size model.
The input sequence length of the visual token embeddings is set to 128.

\vspace{0.15cm}
\noindent\textbf{Other Details.} Following \citep{biten2022latr,powalski2021going}, we use Amazon Textract\footnote{\url{https://aws.amazon.com/textract/}}, Amazon Text-in-Image\footnote{\url{https://docs.aws.amazon.com/rekognition/latest/dg/text-detecting-text-procedure.html}}, and Rosetta \citep{borisyuk2018rosetta} to extract OCR results for document images (\ie, IDL and DocVQA images), non-document images (except for OCR-VQA images), and OCR-VQA images, respectively.
Our implementations are based on the PyTorch \citep{paszke2019pytorch} deep learning framework and the HuggingFace \citep{wolf2020transformers} library.
All experiments are ran on eight NVIDIA A100 GPUs with \verb|cuda11.x|.

% \section{More Ablation Studies}
\section{Information Leak Analyses on OCR-VQA}
\label{sec_supp::infoleak}

In our initial experiments on OCR-VQA, we get accuracy 77.5\% using the MQMA base size model (\vs 69.9\% of the SQSA base size model) on the validation set when we answer 5 questions at a time.
To verify where such big accuracy improvements are from, we conduct detailed analyses on the OCR-VQA dataset.

Unlike other datasets in which questions of the same image are not strongly correlated,
there are correlations among different questions in the OCR-VQA dataset.
For most images in OCR-VQA, the five questions below are asked\\
\emph{Q1: Who wrote this book? / Who is the author of this book?\\
Q2: What is the title of this book?\\
Q3: What type of book is this? / What is the genre of this book?\\
Q4: Is this book related to xxx? / Is this a xxx book?\\
Q5: Is this book related to xxx? / Is this a xxx book?\\}
For Q4 and Q5, one of them has answer ``yes'' and one of them has answer ``no''.
We can see there are correlations among different questions. For example, the title (for Q2) and the type/genre (for Q3) are correlated to each other.
Our MQMA approach can leverage this correlation to improve accuracy.

However, there could be potential information leak from the questions of Q4 and Q5 to the answer of Q3, see the example below.\\
\emph{Q3: What is the genre of this book? - A: religion \& spirituality\\
Q4: Is this book related to religion \& spirituality? - A: yes\\
Q5: Is this book related to computers \& technology? - A: no\\}
As we can see, the question of Q4 contains the answer of Q3.
In addition, if we evaluate the accuracy of Q3 only and other questions,
MQMA gives accuracy 94.0\% for Q3 only and 73.2\% for other questions,
whereas SQSA gives accuracy for 67.0\% for Q3 only and 70.7\% for other questions.
These results show that the MQMA might take information from Q4 or Q5 to answer Q3, \ie, there might be information leak.

To further analyze the information leak issue, we conduct experiments under three settings as follows.
Here we use the MQMA model trained with $n = 5$ for the experiments and we do not add any constraints during training.\\
\textbf{Setting 1:} Answer Q1, Q2, Q4, Q5 together and answer Q3 alone.\\
\textbf{Setting 2.} Answer Q1, Q2, Q3 together and answer Q4, Q5 together.\\
\textbf{Setting 3.} Answer Q1, Q2, Q3 together, answer Q4 alone, and answer Q5 alone.\\
Both of these settings give accuracy 71.5\%, which further confirms answering Q3, Q4, and Q5 together would result in information leak from the questions of Q4 and Q5 to the answer of Q3.
In addition, answering Q4 and Q5 together or alone (Setting 2 and Setting 3) gives the same accuracy, which shows our MQMA approach does not take dataset-specific prior knowledge that there will be one ``yes'' answer and one ``no'' answer for Q4 and Q5.
This is because during training, we do random sampling and ordering, so the training samples could have different numbers of ``yes'' answers and different numbers of ``no'' answers.

To avoid such information leak,
we check the whole dataset and make sure all questions that could result in information leak will not be answered together during both training and testing,
\eg, for the five questions discussed before, we always ensure that Q1, Q2, and Q3 can only be answered together with each other, and Q4 and Q5 can only be answered together with each other.
After doing this, we get accuracy 71.9\% on the OCR-VQA validation set if we answer $n = 5$ questions at a time.

\begin{table}[t]
\centering
% \resizebox{\linewidth}{!}{
\scalebox{0.9}{
\begin{tabular}{l|c|c}
%   \hline
  \toprule
  Approach & \# Questions & Accuracy (\%) \\
  \midrule
  SQSA$_{\text{base}}$ & 1 & 69.7 \\
  \midrule
  MQMA$_{\text{base}}$ & 1 & 70.3 \\
  MQMA$_{\text{base}}$ & 2 & 71.7 \\
  MQMA$_{\text{base}}$ & 3 & 71.9 \\
  MQMA$_{\text{base}}$ & 4 & 71.9 \\
  MQMA$_{\text{base}}$ & 5 & \textbf{71.9} \\
  \bottomrule
\end{tabular}
}
\caption{\textbf{MQMA Ablations:} The influence of the number of questions we answer at a time for MQMA on the OCR-VQA \citep{mishra2019ocr} validation set.}
\label{table_supp::ocrvqa}
\end{table}

\begin{table}[t]
\centering
\resizebox{\linewidth}{!}{
\begin{tabular}{l|c|c|c}
%   \hline
  \toprule
  Approach & \# Questions & TextVQA Accuracy (\%) & ST-VQA ANLS (\%) \\
  \midrule
  SQSA$_{\text{base}}$ & 1 & 60.4 & 68.0 \\
  MQMA$_{\text{base}}$ & 1 & 61.7 & 68.7 \\
  MQMA$_{\text{base}}$ & 2 & \textbf{61.9} & \textbf{69.2} \\
  \bottomrule
\end{tabular}
}
\caption{\textbf{MQMA Ablations:} The influence of the number of questions we answer at a time for MQMA on the TextVQA \citep{singh2019towards} and ST-VQA \citep{biten2019scene} validation set.}
\label{table_supp::textvqa}
\end{table}

\begin{figure*}[t]
    \centering
    \includegraphics[width=\linewidth]{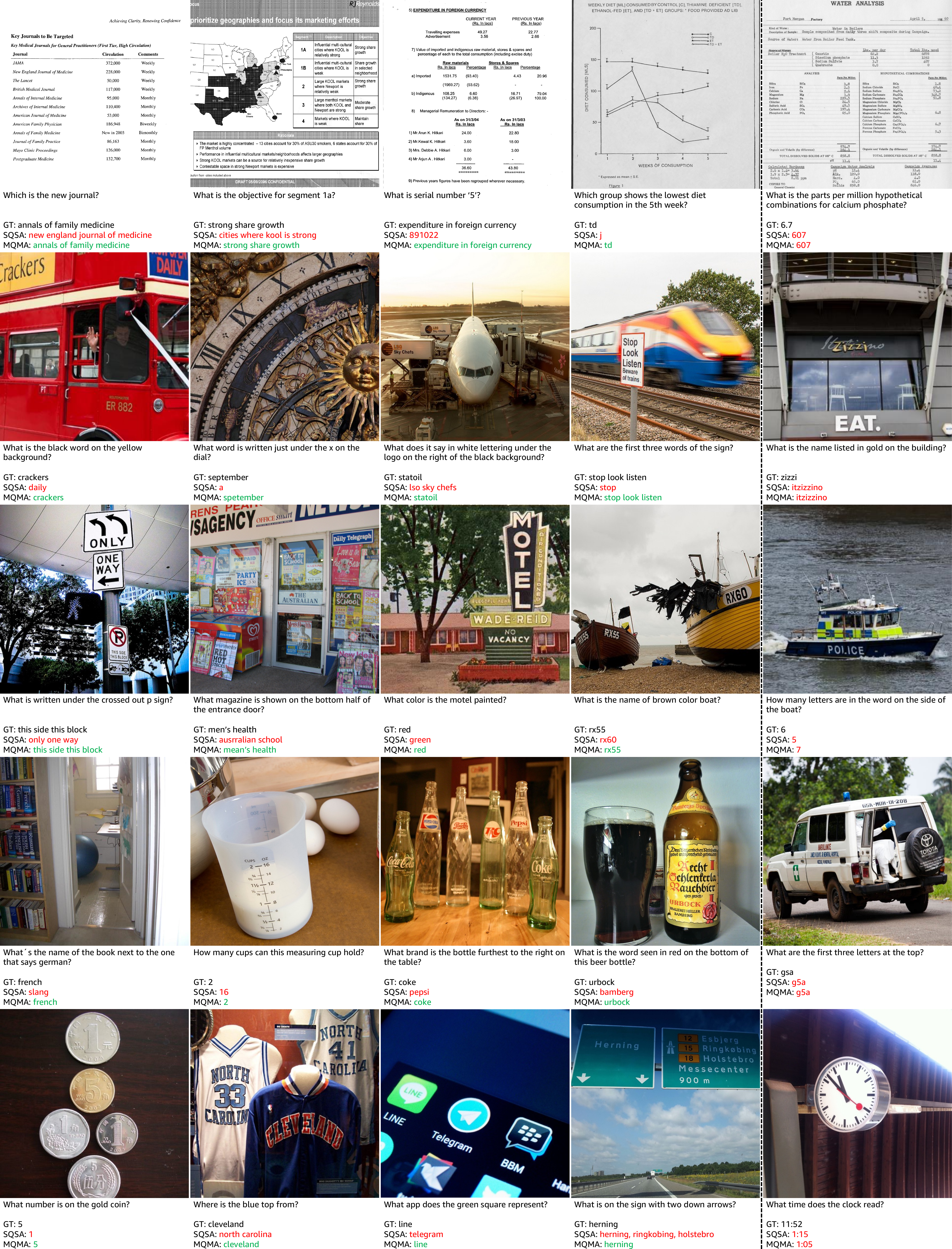}
    \caption{\textbf{Qualitative Comparisons between MQMA and SQSA:}
    The first four columns show examples that MQMA gives correct answers but SQSA gives wrong answers. The last column shows examples that both MQMA and SQSA give wrong answers.
    MQMA shows better multi-modal understanding ability than SQSA. Zoom in to see better.}
    \label{fig_supp::vis_more}
\end{figure*}

\section{The Influence of the Number of Questions on Other Datasets}
\label{sec_supp::n_q}

In our main paper, we only show MQMA results of answering 5 questions at a time on OCR-VQA and results of answering 2 questions at a time on TextVQA and ST-VQA.
Here we should the influence of the number of questions on OCR-VQA, TextVQA, and ST-VQA datasets.
Without loss of generality, we use the base size model and train/test our MQMA approach on the training/validation set.

\vspace{0.15cm}
\noindent\textbf{OCR-VQA.}
Table \ref{table_supp::ocrvqa} shows results of answering different numbers of questions at a time for MQMA on the OCR-VQA \citep{mishra2019ocr} validation set.
Images in OCR-VQA have on average $\sim 5$ questions/image, so we compare results of answering $n = 1$ to $n = 5$ questions at a time.
As we can see, answering different numbers of questions at a time (when $n > 1$) gives very similar accuracy on the OCR-VQA validation set.
Answering $n = 5$ questions at a time gives the highest accuracy on the OCR-VQA validation set, so we only report results of $n = 5$ in our main paper.
Answering $n > 1$ questions at a time gives much higher accuracy than answering $n = 1$ question at a time.
This is because the questions in the OCR-VQA dataset have correlations.
Our MQMA approach can leverage correlations between multiple questions and content to improve accuracy.
Even answering $n = 1$ question at a time for MQMA gives higher accuracy than SQSA, because our MQMA denoising pre-training task aligns the pre-training task and downstream text-VQA task.

\vspace{0.15cm}
\noindent\textbf{TextVQA and ST-VQA.}
Table \ref{table_supp::textvqa} show results of answering different numbers of questions at a time for MQMA on the TextVQA \citep{singh2019towards} and ST-VQA \citep{biten2019scene} validation set.
Here our model is trained on the TextVQA training set only when evaluating on the TextVQA validation set, and is trained on the ST-VQA training set only when evaluating on the ST-VQA validation set.
Images in TextVQA and ST-VQA have only 1 or 2 questions/image,
so we compare results of answering $n = 1$ and $n = 2$ questions at a time.
From the results, we can see answering $n = 2$ questions at a time gives slightly higher numbers than answering $n = 1$ question at a time on TextVQA and ST-VQA, so we only report results of $n = 2$ in our main paper.
Similar to the results on other datasets, even answering $n = 1$ question at a time for MQMA gives higher accuracy than SQSA thanks to the MQMA denoising pre-training task.

\section{Qualitative Results}
\label{sec_supp:qualitative}

We show qualitative results in Figure \ref{fig_supp::vis_more}.
As we can see, our MQMA approach shows better multi-modal understanding ability than SQSA.
There are some failure cases from both MQMA and SQSA. The errors are from multiple aspects, like OCR error and hard images/questions.
For example, for the top right example in Figure \ref{fig_supp::vis_more}, the ground truth is ``6.7'' but both MQMA and SQSA give answer ``607''.
The reason of this wrong prediction is from the OCR error - OCR mis-recognizes the word ``6.7'' as ``607'' and it is hard for models to fix this OCR error.
For the example at the last column of row 3 in Figure \ref{fig_supp::vis_more}, both MQMA and SQSA gives wrong counts for the number of letters in the word ``police''.
Counting is a difficult problem for text-VQA models.
Actually, MQMA gives a reasonable prediction ``7'', because from the appearance of the word in the image it looks like there are ``7'' letters.
There are some cases that even human has difficulty in answering the question -
for the bottom right example, it is hard to answer the time because there is no clear information about which part corresponds to 12 o'clock.

\end{document}